\documentclass{article}
\usepackage{ijcai24}

\usepackage{times}
\usepackage{soul}
\usepackage{url}
\usepackage[hidelinks]{hyperref}
\usepackage[utf8]{inputenc}
\usepackage[small]{caption}
\usepackage{graphicx}
\usepackage{amsmath}
\usepackage{amsthm}
\usepackage{booktabs}
\usepackage{algorithm}
\usepackage{algorithmic}
\usepackage[switch]{lineno}

\usepackage{amssymb,amsmath}
\usepackage{subcaption}
\usepackage{booktabs}
\usepackage{multirow}
\usepackage{pifont}
\usepackage{caption}
\usepackage[labelformat=simple]{subcaption}

\newcommand{\method}[1]{#1}
\newcommand{\cmark}{\ding{51}}%
\newcommand{\xmark}{\ding{55}}%

\title{Contrastive General Graph Matching with Adaptive Augmentation Sampling}

\author{
Jianyuan Bo
\and
Yuan Fang
\affiliations
Singapore Management University, Singapore\\
\emails
\{jybo.2020, yfang\}@smu.edu.sg
}

\begin{document}

\maketitle

\begin{abstract}
Graph matching has important applications in pattern recognition and beyond. Current approaches predominantly adopt supervised learning, demanding extensive labeled data which can be limited or costly. Meanwhile, self-supervised learning methods for graph matching often require additional side information such as extra categorical information and input features, limiting their application to the general case. Moreover, designing the optimal graph augmentations for self-supervised graph matching presents another challenge to ensure robustness and efficacy. To address these issues, we introduce a novel Graph-centric Contrastive framework for Graph Matching (GCGM), capitalizing on a vast pool of graph augmentations for contrastive learning, yet without needing any side information. Given the variety of augmentation choices, we further introduce a Boosting-inspired Adaptive Augmentation Sampler (BiAS), which adaptively selects more challenging augmentations tailored for graph matching. Through various experiments, our GCGM surpasses state-of-the-art self-supervised methods across various datasets, marking a significant step toward more effective, efficient and general graph matching.
\end{abstract}

\section{Introduction}

Graph Matching (GM) refers to the process of \textit{establishing a correspondence between the nodes of two graphs} based on their structure and node attributes \cite{yan2016short,zanfirDeepLearningGraph2018}, and entails various applications. In pattern recognition \cite{shokoufandeh2012many}, graph matching facilitates the alignment of visual patterns within images \cite{wang2018multi}. In bioinformatics \cite{krissinel2004secondary}, it is crucial for understanding protein-protein interaction networks \cite{zaslavskiy2009global}. In social networks, aligning user profiles and their connections is essential for understanding social behaviors \cite{zhang2018mego2vec}.

Despite the widespread applications of graph matching, prevailing methods such as BBGM \cite{bbgm} and NGM \cite{ngm} heavily depend on the supervised learning paradigm, which requires a large number of label to annotate the node correspondence across graphs.
To avoid the costly labeling effort, self-supervised learning (SSL) on graphs \cite{peng2020_graphical_mutual_information_maximization,zeng2021_contrastive_learning_for_graph_classification} becomes a practical alternative. Different from supervised learning, SSL exploits label-free graphs to model intrinsic graph properties general to an application domain, which guides the learning of node representations amenable to graph matching. While label-free graphs are often abundant or easy to obtain, existing SSL-based approaches for graph matching often assume the availability of additional side information besides graphs. On the one hand, approaches like GANN \cite{gann-gm} and IA-SSGM \cite{ia-ssgm} still require graph pairs that are potentially matchable as training data, although they do not need exact annotations on node correspondence. The potentially matchable graph pairs are typically sampled from the same category, under the assumption that graph pairs from distinct categories involve totally different kinds of nodes and cannot be matched. Hence, such approaches require extra categorical information, which can be considered as a form of weak supervision. On the other hand, SCGM \cite{scgm} requires raw image as input in order to perform augmentation, leveraging additional image features in visual applications, and thus cannot be generalized to graph matching in non-visual domains. Additionally, image features may not be available for privacy or commercial considerations.

Thus, toward more general graph matching, an immediate challenge arises: \textit{Is it feasible to employ SSL for graph matching without relying on any side information?} To address this, we develop a \textbf{G}raph-centric \textbf{C}ontrastive framework for \textbf{G}raph \textbf{M}atching (GCGM), using a \emph{large pool} of graph augmentations to compensate for the lack of any side information. As a form of SSL, contrastive learning (CL) \cite{wu2018_instance_discrimination,chen2020simple} regards corresponding/non-corresponding nodes from two different views of the same graph as positive/negative pairs, where a graph can be augmented in different ways to generate different views \cite{hassani2020_contrastive_multiview_representation}. For robustness, we consider both structure- and feature-space augmentations. First, the structure-space augmentations include edge removal, node dropping, and node replacement, etc., which alter the structure of the graph, thereby improving matching against structure variations. Second, the feature-space augmentations include feature masking and scaling, which manipulate nodes features, thereby improving matching against feature variations. Generally speaking, each type of augmentation focuses on a specific kind of variation across graphs to enhance a specific aspect of matching capability. Moreover, each type of augmentation (e.g., node dropping) in fact entails a family of augmentations defined by some hyperparameters (e.g., probability of dropping a node) to vary the difficulty levels within each type. Together, we form a large, comprehensive pool of augmentations to improve the robustness of graph matching.

However, given a comprehensive pool of graph augmentations, a second challenge follows: \textit{How do we select the most effective augmentations for graph matching?} The pool of graph augmentations includes various types of structure- and feature-space augmentations, and each type can instantiate a family of augmented views by tuning the hyperparameters of that type.  In existing graph contrastive learning, hyperparameter tuning for each type of augmentation is the prevailing solution \cite{youGraphContrastiveLearning2020}, which can be expensive and tends to overfit. A few automated methods exist in selecting the right type of augmentations. JOAO \cite{joao} always exploits the most challenging augmentation w.r.t.~the current loss, while recent research \cite{cai2020_negatives_equal,robinson2021_CONTRASTIVE_HARD_NEGATIVE_SAMPLES,wang2021_understanding_contrastive_behaviour} finds that the hardest instances are not necessarily the best for learning.  
Meanwhile, GCA \cite{gca} adaptively augments each graph through edge removal based on node centrality and masking unimportant node features, which requires custom design for each type of augmentation and does not generalize to other types. To counter this challenge, we present a \textbf{B}oosting-\textbf{i}nspired Adaptive \textbf{A}ugmentation \textbf{S}ampler (BiAS) that automatically and adaptively samples graph augmentations beyond na\"ive uniform sampling, providing a general solution to the otherwise expensive task of selecting and tuning the optimal augmentations.

By addressing the challenges, we make three major contributions to graph matching. 
(1) We propose GCGM, a graph-centric contrastive framework for general graph matching, based on a large pool of graph augmentations without the need of any labeled data or side information. 
(2) We introduce BiAS, an adaptive augmentation selection strategy for GCGM, to optimize graph matching and eliminate manual tuning effort over the pool of augmentations. 
(3) We conduct experiments on both real-world and synthetic datasets, demonstrating that our approach consistently outperforms state-of-the-art self-supervised baselines.

\section{Related Work}
\paragraph{Deep Graph Matching}
In supervised graph matching, GMN \cite{zanfirDeepLearningGraph2018} has pioneered the use of deep learning for GM. Subsequent research, like PCA \cite{wang2020combinatorial}, has advanced this by learning node, edge, and affinity features end-to-end. Another notable approach is the channel-independent embedding with Hungarian attention in deep graph matching \cite{cie}.

Furthermore, NGM \cite{ngm} extends GM to hypergraphs and multi-graph matching (MGM). Additionally, BBGM \cite{bbgm} addresses GM using differentiable operations on combinatorial solvers. Besides, addressing outliers and noise is crucial for obtaining accurate matches. Notably, RGM \cite{rgm} introduced a reinforcement learning approach with a revocable action scheme to counteract the impact of outliers, using a pre-trained BBGM as a feature extractor that still requires labels.

While supervised graph matching methods have progressed using large labeled datasets, unsupervised or self-supervised approaches have emerged to tackle label scarcity. GANN utilizes graduated assignment mechanisms that allows for joint MGM and clustering. SCGM introduces a two-stage contrastive learning framework where image augmentation is essential, necessitating visual access and limiting its use in other domains. Given the advantage of self-supervised learning, our research primarily benchmarks against these approaches in the evolving landscape of graph matching.

\paragraph{Graph Contrastive Learning}
Graph learning has advanced significantly with the advent of graph neural networks (GNNs) \cite{velivckovic2017graph} and the introduction of contrastive learning methods \cite{chen2020simple}, leading to the novel concept of graph contrastive learning (GCL) \cite{youGraphContrastiveLearning2020}. As a major branch of self-supervised learning, GCL enhances the representation of nodes \cite{tan2022_supervised_Graph_Contrastive_Learning_for_Few_Shot_Node_Classification}, edges \cite{li2023s}, and entire graphs \cite{peng2020_graphical_mutual_information_maximization} by contrasting corrupted views augmented by different graph augmentations \cite{xu2022_Contrastive_cascade_graph_learning}, providing superior performance to traditional graph learning techniques. However, these approaches face the challenge of designing effective augmentations, which may involve custom designs or extensive hyperparameter tuning \cite{youGraphContrastiveLearning2020,zhao2021_data_augmentation_for_graph_neural_networks}, 
posing a significant overhead.

\begin{figure*}[t]
   \centering
   \includegraphics[width=0.95\textwidth]{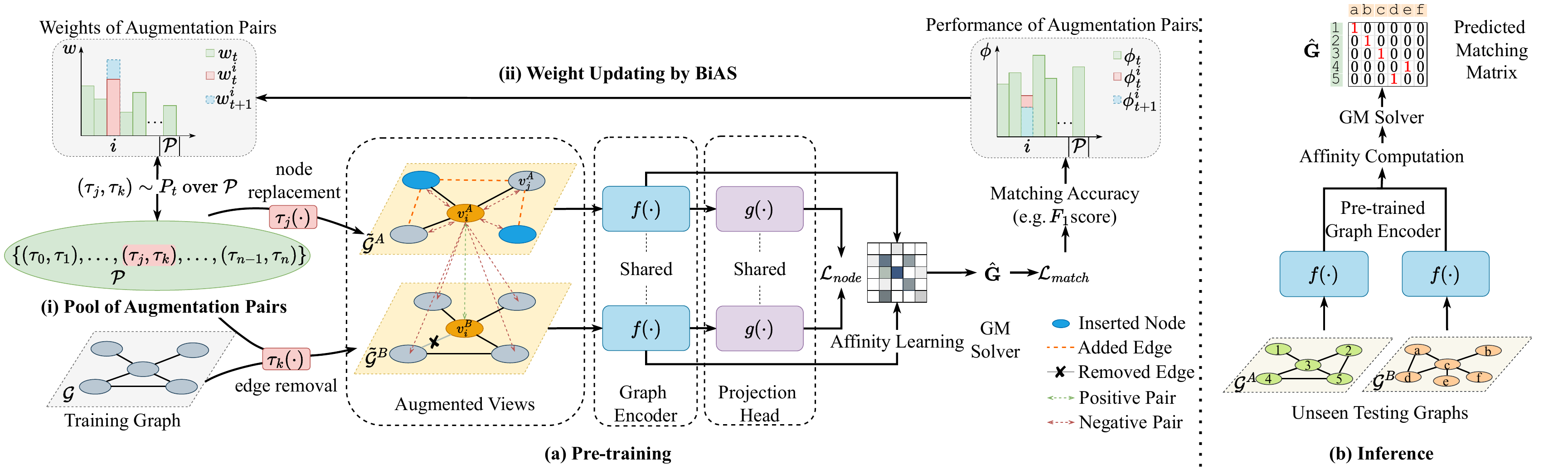}
   \caption{Framework of the proposed GCGM with BiAS. (a) Pre-training: Each graph $\mathcal{G}$ in the training set is augmented by sampling augmentation pairs from a large pool, guided by the BiAS strategy. (To be clear, we only show one training graph here; the same augmentation process is applied to every graph in the training set.) (b) Inference: The pre-trained model is frozen and applied to unseen graph pairs.}
   \label{fig:gcgm}
\end{figure*}

\section{Proposed Method}
In this section, we introduce our GCGM framework, starting with the overall framework and preliminaries, followed by our augmentation designs and adaptive sampling strategy.

\subsection{Overall Framework}

\paragraph{Pre-training} Fig.~\ref{fig:gcgm}(a) depicts the self-supervised pre-training phase. The illustrated augmentation process is applied across all graphs in the entire training set. 

Specifically, for each graph $\mathcal{G} = \{\mathbf{X}, \mathbf{A}\}$ in a batch during pre-training, defined by node features $\mathbf{X} \in \mathbb{R}^{N \times F}$ and the adjacency matrix $\mathbf{A}$, we sample a augmentation pair $(\tau_j, \tau_k)$ based on their weights w.r.t.~the current mini-batch, from a randomly initialized augmentation pairs pool as shown in Fig.~\ref{fig:gcgm}(a)(i).
Through graph augmentation, we produce two distinct views of each input graph: $\tilde{\mathcal{G}}^A$ and $\tilde{\mathcal{G}}^B$, with $N_A$ and $N_B$ nodes respectively, which will then be processed by an graph encoder, capturing both local and global graph patterns. Node embeddings are then refined through the node-level contrastive strategy, optimizing the embedding by matching congruent nodes across views and differentiating non-correspondence nodes. Finally, the encoded embeddings will be processed by the subsequent affinity learning module and GM solver to produce the predicted matching matrix $\hat{\mathbf{G}} \in \mathbb{R}^{N_A \times N_B}$.

Enhancing this workflow, as illustrated in Fig.~\ref{fig:gcgm}(a)(ii), our BiAS technique dynamically updates the weights of the augmentation pair according to the matching accuracy between the two corresponding augmented views. This adjustment influences the sampling probability for subsequent mini-batches. By doing so, BiAS adaptively samples challenging augmentations tailored for GM, mitigating the necessity for exhaustive hyperparameter tuning and augmentation pool selection. 

\paragraph{Inference}
In the inference phase, as shown in Fig.~\ref{fig:gcgm}(b), we apply the pre-trained, now-frozen graph encoder and affinity learning module to unseen graph pairs. The GM solver predicts the matching matrix, which determines the node-to-node alignment for each input graph pair.

We will next introduce some technical preliminaries (Sect.~\ref{sec:model:prelim}), before discussing our main contributions on the comprehensive pool of graph augmentations (Sect.~\ref{sec:comprehensive_graph_augmentations}) and our adaptive sampler BiAS (Sect.~\ref{sec:bias}), respectively.

\subsection{Preliminaries}\label{sec:model:prelim}
\paragraph{Graph Encoder} During training, our proposed graph encoder takes an augmented graph $\tilde{\mathcal{G}}$ as input to produce both node and graph-level representations as output. Given the comprehensive graph augmentations we carry out, it is essential to leverage information at multiple granularities to preserve both local and global graph patterns. To achieve this, we have opted for the commonly used GraphSAGE framework \cite{hamilton2017inductive} as our base graph encoder, denoted as $f(\cdot)$ in Fig.~\ref{fig:gcgm}. This encoder captures both local node-level embedding $\tilde{\mathbf{h}}_v$ using GraphSAGE layers with skip connections, and global graph-level embedding $\tilde{\mathbf{h}}_{\tilde{\mathcal{G}}}$ via a readout function. The node- and graph-level embeddings are then utilized for the affinity learning and GM solver modules, to be elaborated next. For detailed specifications of the encoding function $f(\cdot)$, refer to Appendix A.

\paragraph{Affinity Learning and GM Solver} On one hand, the affinity learning module accepts these embeddings from the two views (graphs) as input and generates an affinity matrix, containing the pairwise affinity scores between the nodes from the two views. Subsequently, the GM solver leverages the affinity matrix to determine the node correspondence across the two graphs. We will use the affinity learning module and GM solver following previous work \cite{bbgm,ngm}.

\paragraph{Contrastive Loss} After processing the two augmented views of the graph through our encoder, as depicted in Fig.~\ref{fig:gcgm}(a), we employ a contrastive framework to learn node embeddings. Adapting from GRACE \cite{zhuDeepGraphContrastive2020}, we employ both intra-view and inter-view contrastive objectives to enhance the comparison and alignment of nodes between distinct graph views.

First, the intra-view contrastive objective aims to differentiate anchor nodes from other nodes within the same view by maximizing their dissimilarity. Specifically, in view $\tilde{\mathcal{G}}^A$ with $N_A$ nodes, the dissimilarity between the anchor node ${v}^A_i$ with feature $\tilde{\mathbf{h}}_{i}^{A}$ and the negative nodes is
\begin{equation}
    \textstyle
	\ell^A_\text{intra}(v_i)=\sum_{j=1; j \neq i}^{N_A} \exp \left(\operatorname{sim}\left(g(\tilde{\mathbf{h}}_{i}^{A}), g(\tilde{\mathbf{h}}_{j}^{A})\right) / T \right),
\end{equation}
where $\operatorname{sim}(\cdot,\cdot)$ denotes cosine similarity, $g(\cdot): \mathbb{R}^{d_h} \to \mathbb{R}^{d'_h}$ is a non-linear projection head, and $T$ is the temperature hyperparameter. This approach ensures that  distinct nodes within the same view become dissimilar.

Second, our inter-view contrastive objective, operating across distinct augmented views $\tilde{\mathcal{G}}^A$ and $\tilde{\mathcal{G}}^B$, maximizes the similarity of corresponding nodes across views while minimizing that of an anchor node and all other nodes in the opposite view. We first sum up the similarities across all inter-view pairs, centered around the anchor node in view $\tilde{\mathcal{G}}^A$:
\begin{equation}
    \textstyle
	\ell^A_\text{inter}(v_i) =\sum_{j=1}^{N_B} \exp \left(\operatorname{sim}\left(g(\tilde{\mathbf{h}}_{i}^{A}), g(\tilde{\mathbf{h}}_{j}^{B})\right) / T \right).
\end{equation}

Integrating both intra- and inter-view objectives based on the NT-Xent loss \cite{chen2020simple}, the  contrastive loss for an anchor node is
\begin{equation}
    \textstyle
	\ell^A_{i} = -\log \frac{\exp \left(\operatorname{sim}\left(g(\tilde{\mathbf{h}}_{i}^{A}), g(\tilde{\mathbf{h}}_{i}^{B})\right) / T\right)}{\ell^A_\text{intra}(v_i) + \ell^A_\text{inter}(v_i)}.
\end{equation}
The total contrastive loss averages across all anchor nodes in the two views: 
\begin{align}
  \textstyle
  \mathcal{L}_{\text {node }}=\frac{1}{N_A+N_B}\left(\sum_{i=1}^{N_A} \ell_i^A+\sum_{i=1}^{N_B} \ell_i^B\right).
\end{align}
In essence, our contrastive strategy attempts to align corresponding nodes in the two views to achieve graph matching.

\paragraph{Graph Matching Loss} While our contrastive loss refines individual node representations, our approach recognizes the importance of holistic graph matching. To this end, we incorporate a graph matching loss $\mathcal{L}_{\text {match}}$. This loss bridges the gap between our predicted matching matrix, $\hat{\mathbf{G}}$, and a self-labeled ground-truth matching matrix $\mathbf{G}^\text{self}$. $\mathbf{G}^\text{self}$ can be readily determined without supervision, as $\tilde{\mathcal{G}}^A$ and $\tilde{\mathcal{G}}^B$ originate from the same input graph. 
\begin{equation}
	\mathcal{L}_{\text {match}}=\mathcal{L}\left(\hat{\mathbf{G}}, \mathbf{G}^\text{self}\right)
 \end{equation}
The specific GM solver used determines the form of the loss: a permutation loss for NGM \cite{ngm}, or a Hamming loss for BBGM \cite{bbgm}.

Our overall loss $\mathcal{L} = \mathcal{L}_{\text {node}} +  \mathcal{L}_{\text {match}}$ combines contrastive loss and graph matching loss, ensuring that the overall graph representations are optimized for graph matching tasks.

\begin{figure}[!t]
    \centering
    \begin{subfigure}{0.1\textwidth}
        \centering
        \includegraphics[width=0.75\linewidth]{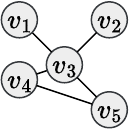}
        \caption{}
        \label{fig:aug:input_graph}
    \end{subfigure}
    \begin{subfigure}{0.1\textwidth}
        \centering
        \includegraphics[width=0.75\linewidth]{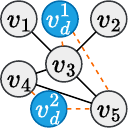}
        \caption{}
        \label{fig:aug:node_insertion}
    \end{subfigure}
    \begin{subfigure}{0.1\textwidth}
        \centering
        \includegraphics[width=0.75\linewidth]{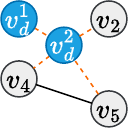}
        \caption{}
        {\label{fig:aug:node_replacement}}
    \end{subfigure}
    \begin{subfigure}{0.1\textwidth}
        \centering
        \includegraphics[width=0.75\linewidth]{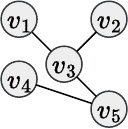}
        \caption{}
        {\label{fig:aug:edge_removal}}
    \end{subfigure}
    \caption{Graph augmentation: \textbf{(a)} input graph; \textbf{(b)} node insertion; \textbf{(c)} node replacement; \textbf{(d)} edge removal. The blue node represents the inserted node, and the dotted edge indicates the added edge.}
    \label{fig:aug}
\end{figure}

\subsection{Comprehensive Graph Augmentations}
\label{sec:comprehensive_graph_augmentations}
Appropriate graph augmentations are crucial for the success of contrastive learning toward general graph matching. 
As illustrated in Fig.~\ref{fig:gcgm}(a), we propose to select from a comprehensive pool of augmentation pairs to generate two views for each input graph.
While existing methods often adopt a small number of standard augmentations, we argue that, in the absence of side information for graph matching, we need to formulate a comprehensive pool of augmentations for robustness. In particular, we not only focus on a diverse range of augmentation types to improve different aspects of matching, but also expand each augmentation type into a family of augmentations with varying difficulty levels.
In the following, we first introduce different types of augmentations; then, we discuss the construction of the pool of paired augmentations for contrastive learning. 

\paragraph{Types of Augmentation} We introduce four major graph augmentation types to robustly handle real-world matching scenarios, most notably the complexities in managing outliers, noises \cite{cho2014finding} and high-order structures. The first three types operate in the structure space, while the last one operates in the feature space. 

\emph{Node Insertion (NI).}
Recognizing outliers, or nodes that are exclusive to only one of the graph in a candidate matching pair, is a crucial aspect for graph matching. 
Inspired by node addition in contrastive learning for graph classification \cite{zeng2021_contrastive_learning_for_graph_classification}, we propose a node insertion strategy to simulate outliers. As shown in Fig.~\ref{fig:aug:node_insertion}, given a randomly selected subset of nodes from the original input graph, we aggregate their features to construct a dummy node (e.g., $v_d^1$) and randomly link it to randomly chosen nodes not in the given subset. The augmented graph view with the dummy nodes can provide contrastive samples to learn outliers.

\emph{Node Replacement (NR).}
This is a variant of node insertion, to simulate the matching scenario where there can still be outliers even though the total number of nodes are the same in a candidate pair. As illustrated in Fig.~\ref{fig:aug:node_replacement}, we first drop some nodes randomly from the original graph (e.g., $v_1$, $v_3$) along with their incident edges, and then insert an equal number of dummy nodes (e.g., $v_d^1$, $v_d^2$) following the node insertion strategy above.

\emph{Edge Removal (ER).} In many real-world graphs, first-order connections (i.e., edges to one-hop neighbors) are often noisy or sparse. This limitation can be alleviated by considering high-order structural connectivity or node features. 
To recognize node correspondences without overly relying on first-order connections, we adopt an edge removal strategy \cite{hassani2020_contrastive_multiview_representation}. As shown in Fig.~\ref{fig:aug:edge_removal}, a subset of edges (e.g., $v_3$--$v_4$) is randomly removed from the graph, which encourages the model to focus more on the high-order connections  (e.g., $v_3$--$v_5$--$v_4$) or invariant features.

\emph{Feature Scaling (FS).}
Node features in graphs can exhibit significant variability and contain considerable noises. To address this, we present two feature scaling techniques \cite{laskin2020reinforcement}. First, node features are scaled using a univariate random variable, such that all dimensions of the feature vector of the same node are scaled by the same factor. Second, we apply a multivariate random variable, such that different scaling factors can be applied to different dimensions. 

In addition to above four major types of augmentations designed to deal with different matching scenarios, we further adopt a series of standard graph augmentations aimed to capture general graph properties that are widely used in previous work. We summarize the four major of augmentations used in our approach in Tab.~\ref{table:major_augmentations}; the additional types of augmentation are in Appendix B. Note that each type of augmentation is also associated with one or more hyperparameters, adjusting which can instantiate different augmentations with variable difficulty.  
Additionally, we introduce a special type $\tau_{\varnothing}$ that does nothing, which enables us to generate a view using the the original graph.

Formally, let $\mathcal{T}$ denote the set of augmentation types including $\tau_{\varnothing}$. Further define  $\mathcal{I}_\tau$ as the set of augmentations that can be instantiated from type $\tau\in \mathcal{T}$, by sampling different hyperparameter values associated with $\tau$. Hence, the full set of augmentations under consideration is $\mathcal{I}=\cup_{\tau\in\mathcal{T}} \mathcal{I}_\tau$.

\begin{table}[tbp]
     \centering
     \addtolength{\tabcolsep}{-1.2mm}
     \resizebox{\linewidth}{!}{
     \begin{tabular}{@{}c|c|cc@{}}
     \toprule
     Type & Matching scenario & Hyperparameters\\
     \midrule\midrule
     \multirow{4}*{NI} & \multirow{4}*[0.5ex]{\shortstack[c]{node outlier\\(unequal node count\\in two views)}} &\multirow{4}*[0.5ex]{\shortstack[c]{$p_\text{ni} \in [0.1, 0.9]$: fraction of nodes inserted;\\ $k_\text{ni}\geq 2$: size of subset; \\ $aggr_\text{ni}\in \{\text{mean}, \text{max}\}$: aggregation function; \\ $e_\text{ni} \geq 1$: number of edges inserted}}\\
     & & & \\
     & & & \\
    & & & \\\midrule
     \multirow{4}*{NR} & \multirow{4}*[0.5ex]{\shortstack[c]{node outlier\\(equal node count\\in two views)}} & \multirow{4}*[0.5ex]{\shortstack[c]{$p_\text{nr} \in [0.1, 0.9]$: fraction of nodes replaced; \\ $k_\text{nr} \geq 2$: size of subset; \\ $aggr_\text{nr}\in \{\text{mean}, \text{max}\}$: aggregation function; \\ $e_\text{nr}\geq 1$: number of edges inserted}}\\
     & & &\\
     & & &\\
     & & &\\
     \midrule
     \multirow{2}*{ER} & \multirow{2}*{\shortstack[c]{sparse/noisy first-\\ order connections}} & \multirow{2}*{$p_\text{er} \in [0.1, 0.9]$: probability of each edge being removed}\\
     & & &\\
     \midrule
     \multirow{2}*{FS} & \multirow{2}*{\shortstack[c]{feature variations \\ \& noises}} & \multirow{2}*{\shortstack[l]{$\alpha \in [0.2, 0.8]$: lower bound of uniform distribution;\\ $\beta\in [1.2, 1.8]$: upper bound of uniform distribution}}\\
     & & &\\
     \bottomrule
     \end{tabular}}
     \caption{Details of the four major types of graph augmentation.}
     \label{table:major_augmentations}
 \end{table}

\paragraph{Pool of Paired Augmentations} As illustrated in Fig.~\ref{fig:gcgm}(a)(i), given an input graph, we generate two augmented views of the graph by applying a pair of augmentations.
Existing graph contrastive learning approaches mostly employ well-tuned hyperparameters for every augmentation type. However, given a large number of augmentation types and associated hyperparameters, extensive tuning can be combinatorially costly and tends to overfit. Moreover, recent research shows \cite{cai2020_negatives_equal} that training can often benefit from samples of variable difficulty level. In our case, as the difficulty of augmentation is controlled by its associated hyperparameters, it is not ideal to only rely on one set of tuned hyperparameters. 

In our approach, given an input graph, we propose to apply a pair of augmentations sampled from $\mathcal{P}$, the pool of all possible pairs. Most generally, we can define $\mathcal{P}=\mathcal{I}^2$, i.e., the Cartesian product of $\mathcal{I}$ with itself, where $\mathcal{I}$ is the set of all augmentations. However, we exclude two special cases: (1) Both augmentations in the pair belong to the $\tau_{\varnothing}$ type, since both views being the original graph does not form a meaningful contrastive example; (2) both belong to the mixup type following earlier work \cite{scgm} which tends to give poor empirical performance.

\subsection{Sampling Augmentation Pairs via BiAS}
\label{sec:bias}
Given a pool of augmentation pairs $\mathcal{P}$, we sample a pair to be applied to each input graph to form the contrastive example for training. Adopting a uniform sampler is the most straightforward approach, but its disadvantage is twofold. First, it does not distinguish the importance of different augmentations; second, it is static and task irrelevant, which does not adapt to the fluctuations in training.

In our GCGM framework, we propose BiAS, a \textbf{B}oosting-\textbf{i}npired Adaptive \textbf{A}ugmentation \textbf{S}ampler. As the name suggests, BiAS is adaptive to training fluctuations, inspired by the boosting technique \cite{schapire1999_brief_boosting} in ensemble learning that iteratively refines weak learners by focusing on more difficult examples. 
Specifically, augmentation pairs that result in subpar matching performance in a mini-batch during training are marked as difficult, suggesting that the model needs further adaptation to data variations introduced by such augmentations. These challenging augmentation pairs subsequently receive more frequent sampling by increasing their probability. Conversely, pairs that can be solved effortlessly with high accuracy are deemed easy and are less frequently sampled in the following mini-batches. Essentially, this strategy biases toward certain augmentations dynamically, addressing the drawbacks of a uniform sampler.

Formally, we introduce a weight updating scheme to dynamically adjust the sampling probability of each augmentation pair from the pool $\mathcal{P}$, as illustrated in Fig.~\ref{fig:gcgm}(a)(ii).
Specifically, each augmentation pair $i\in \mathcal{P}$ is associated with a weight $w_{t}^i$ in the $t$-th  mini-batch. The weight is updated in the next mini-batch as follows. 
\begin{equation}
 w_{t+1}^i = \lambda  w_t^i + (1 - \lambda)  e^{\alpha  (1 - \phi_t^i)},
\end{equation}
where $\phi_t^i$ indicates the mean performance score for all matchings where augmentation pair $i$ was previously applied up to the current mini-batch $t$. In our implementation, we adopt the $F_1$ score as the performance score (see Appendix E), which measures the accuracy of the predicted matching against the contrastive self-supervision generated from the two graph views. The hyperparameter $\alpha \geq 1$ regulates the magnitude of weight adjustment, while $\lambda \in [0,1]$ serves the purpose of momentum update to constrain the potentially exponential increase in weights of those very challenging pairs yielding low $\phi_t^i$  repeatedly. 
The weights in the $t$-th mini-batch are then normalized into a probability distribution over the pool of augmentation pairs $\mathcal{P}$, given by $P_t(i) = {w_{t}^i}/\sum_{j \in \mathcal{P}} w_{t}^j$.
The initial weight $w^i_0$ is set to $e^{\alpha}$ for all augmentation pairs, i.e., we start with a uniform sampler.

Finally, given an input graph in the $t$-th mini-batch, we generate two views using an augmentation pair $i\sim P_t(i)$ drawn from the distribution over the pool $\mathcal{P}$. Pseudocode of our method can be found in Appendix A.

In summary, BiAS offers an adaptive strategy to choose the right augmentations. Compared to the uniform sampling strategy, BiAS dynamically selects the most useful augmentation pairs in each mini-batch during training. Compared to the conventional approach of tuning the hyperparameters of each augmentation type, BiAS is much cheaper and can benefit from a wider range of augmentations.

\section{Experiment}
In this section, we empirically evaluate the proposed model GCGM and BiAS.

\subsection{Experiment Setup}
\paragraph{Datasets} We tested three real-world datasets. (1) Pascal VOC \cite{BourdevICCV09,EveringhamIJCV10} includes images from 20 classes; (2) Willow \cite{ChoICCV13} offers 256 images over five classes; (3) SPair-71k \cite{min2019spair} has 70,958 image pairs across 18 classes. Besides, we followed a recent work \cite{rgm} to generate a synthetic dataset from random 2D node coordinates for the general non-visual domain. All graphs are constructed based on Delaunay triangulation. More dataset details are presented in 
Appendix C.

\paragraph{Baselines} We assessed the performance of GCGM against diverse baselines, including supervised, learning-free, and unsupervised methods. Our emphasis is benchmarking against graph-centric self-supervised methods, given our exclusion of labeled data or side information. Supervised methods included cutting-edge techniques like CIE \cite{cie}, BBGM \cite{bbgm}, and NGMv2 \cite{ngm}. Learning-free methods included RRWM \cite{cho2010reweighted}, IFPF \cite{leordeanu2009integer}, and SM \cite{leordeanu2005spectral}. For self-supervised baselines, we compared against GANN-GM \cite{gann-gm} and SCGM \cite{scgm}.

\paragraph{Settings and Hyperparameters} GCGM with BiAS is a generic framework that can be applied to any GM solver. To demonstrate this, we paired them with both BBGM \cite{bbgm} and NGMv2 \cite{ngm}. As GCGM is graph-centric, we froze the image backbone on the visual datasets, and treated each input image solely as a graph. Our implementation is based on ThinkMatch \cite{ngm}, and we kept the original configurations for both solvers. Our reported results for supervised methods and SCGM might be slightly lower than their original papers due to our 80:20 train-validation split from the original training set, as the original splits lack a validation set.
We repeated the splits five times using varied random seeds. We tuned the hyperparameters of our encoder using Tree-structured Parzen Estimator (TPE) \cite{bergstra2011algorithms} via Optuna \cite{akiba2019optuna} based on the validation set. For BiAS, we set $\lambda = 0.8$, $\alpha=3$, $|\mathcal{P}|=512$. However, for the Willow dataset, due to its smaller size, we adjust $|\mathcal{P}|$ to 128. 
Early stopping was applied if performance improvements were below the threshold $\epsilon = 0.001$. Detailed model and parameter configurations can be found in 
Appendix A.

For other supervised and SSL baselines, we also applied the same early stopping criterion. On visual datasets, we allowed them to fine-tune their visual backbone and employ visual augmentations (if any) per their original paper. More detailed baseline settings can be found in Appendix D.

\begin{table}[tbp]
    \centering
    \small
    \addtolength{\tabcolsep}{-1mm}
    \resizebox{\linewidth}{!}{
    \begin{tabular}{@{}l|ll|l|ll@{}}
    \toprule
    \multirow{2}*{Methods} &\multicolumn{2}{c|}{Pascal VOC} & \multicolumn{1}{c|}{Willow} & \multicolumn{2}{c}{SPair-71k} \\
      & \multicolumn{1}{c}{Intsec}  & \multicolumn{1}{c|}{Unfilt} & \multicolumn{1}{c|}{Intsec} & \multicolumn{1}{c}{Intsec} & \multicolumn{1}{c}{Unfilt} \\
     \midrule\midrule
    \method{CIE} (SUP)  & 66.8$\pm$0.4 & \multicolumn{1}{c|}{-} & 82.6$\pm$0.2& 69.3$\pm$0.3& \multicolumn{1}{c}{-}\\
   \method{BBGM} (SUP)  & 77.3$\pm$0.1 & 55.0$\pm$0.1 & 96.2$\pm$0.1 & 77.7$\pm$0.2 & 48.4$\pm$0.2\\
   \method{NGMv2} (SUP) & 76.8$\pm$0.1& 56.7$\pm$0.1 & 94.5$\pm$0.3 & 76.6$\pm$0.2 &49.8$\pm$0.08\\
    \midrule
    \method{IPFP} & 45.8$\pm$0.02 & 31.5$\pm$0.002& 80.1$\pm$0.06& 57.0$\pm$0.04& 31.7$\pm$0.01\\
   \method{RRWM}  & 47.2$\pm$0.02 &31.7$\pm$0.001& 83.4$\pm$0.09 & 58.6$\pm$0.05 & 32.3$\pm$0.01\\
    \method{SM} & 46.2$\pm$0.03&30.4$\pm$0.002 & 81.3$\pm$0.08& 57.7$\pm$0.04& 30.3$\pm$0.01\\\midrule
    \method{GANN-GM\^} & 34.5$\pm$0.3 & 23.4$\pm$0.2 & 89.3$\pm$0.1& 34.7$\pm$0.4 & 19.4$\pm$0.3\\
    \method{SCGM+BBGM} & 54.8$\pm$0.05& \underline{36.6}$\pm$0.04 & 93.1$\pm$0.08 & 60.2$\pm$0.05 & 34.1$\pm$0.01\\
   \method{SCGM+NGMv2} & 50.8$\pm$0.1& 32.9$\pm$0.03 & 84.2$\pm$0.1 & 59.8$\pm$0.1 & 30.5$\pm$0.3\\\midrule
    \method{GCGM+BBGM} & \underline{56.8}$\pm$0.02 & 36.2$\pm$0.01 & \underline{94.4}$\pm$0.3 & \underline{60.6}$\pm$0.1 & \textbf{35.9}$\pm$0.07\\
    \method{GCGM+NGMv2} & \textbf{57.3}$\pm$0.11 & \textbf{37.4}$\pm$0.07 & \textbf{95.0}$\pm$0.1 & \textbf{62.6}$\pm$0.02 & \underline{35.4}$\pm$0.07\\
    \bottomrule
    \end{tabular}}
    \caption{Performance (\%) on real-world datasets. Supervised methods are marked with `SUP', included for reference only. ${\text-}$: running exception in handling the matching scenario. $\hat{~}$: self-supervised methods that require categorical information. Bold/underlined: best/runner-up results (excluding supervised methods).}
    \label{table:real-world}
\end{table}

\paragraph{Evaluation Metric} We used the $F_1$ score for evaluation (see Appendix E), combining precision and recall for a balanced measure, which is vital for matching graphs with outliers. We repeated the experiments on the five different splits described above, and report their  average and standard deviation in $F_1$. 

\subsection{Performance Evaluation} In our experiments, we adopted the two pairwise GM settings outlined by SCGM. The \textit{Intersection} (Intsec) setting 
confines keypoints (i.e., nodes) to only those shared between the input graph pair. 
Meanwhile, the \textit{Unfiltered} (Unfilt) setting includes all keypoints, accounting for potential outliers and variations in keypoint counts between graphs. Notably, the \textit{Unfiltered} setting is more challenging because of the presence of the outliers. Note that in the case of the Willow dataset, the \textit{Intersection} and \textit{Unfiltered} scenarios are equivalent due to each graph class containing 10 shared keypoints (i.e., nodes) without outliers. Consequently, we only report the \textit{Intersection} results. For training, we used the original graph without any filtering, retaining all the keypoints.

\paragraph{Real-world Datasets} In Tab.~\ref{table:real-world}, we present the performance evaluation on the visual datasets (see detailed per-class results in Appendix F). We make several observations.

First, in terms of overall performance, GCGM consistently surpasses other SSL baselines and learning-free methods. On the Willow dataset, GCGM outperforms even some supervised methods. This result highlights the effectiveness of our contrastive approach, especially given that the baselines are allowed to fine-tune the visual backbone and apply image augmentations (if applicable), whereas our GCGM is graph-centric and does not rely on these visual elements.

Second, when paired with the same GM solver, GCGM consistently outperforms the state-of-the-art SCGM. This finding highlights the flexibility and robustness of GCGM when working with different GM solvers. Specifically, when pairing with NGMv2, a GNN-based solver, our approach tends to perform the best. For brevity, in subsequent results, we default to reporting only the performance of GCGM using the NGMv2 solver. 

Third, GCGM demonstrates solid performance in both
\textit{Intersection} and \textit{Unfiltered} settings. In particular, the \textit{Unfiltered} setting is often more challenging given the presence of outliers. Hence, all methods give lower performance in this setting, while GCGM continues to maintain a clear advantage. 

\paragraph{Synthetic Dataset}
\begin{table}[tbp]
    \centering
    \small
    \resizebox{0.7\linewidth}{!}{
    \begin{tabular}{@{}l|cc@{}}
    \toprule
    \multirow{2}*{Methods} &\multicolumn{2}{c}{Synthetic} \\
     & Intsec & Unfilt \\
     \midrule\midrule
     \method{GANN-GM\^} & 11.2$\pm$0.04 & 10.2$\pm$0.03 \\
     \method{SCGM + BBGM} & 33.5$\pm$2.0 & 24.3$\pm$1.2\\
     \method{SCGM + NGMv2} & 35.2$\pm$0.6 & 25.0$\pm$0.4 \\\midrule
     \method{GCGM} & \textbf{58.1}$\pm$0.5 & \textbf{39.9}$\pm$0.4\\
    \bottomrule
    \end{tabular}}
    \caption{Performance (\%) of SSL methods on the synthetic dataset.}
    \label{table:synthetic}
\end{table}

The performance of various SSL methods on the synthetic dataset is presented in Tab.~\ref{table:synthetic}, providing a focused and fair comparison since our GCGM is self-supervised. 
Note that supervised methods (CIE, BBGM and NGMv2) and learning-free methods (IPFP, RRWM and SM) as shown in Tab.~\ref{table:real-world} are for reference only; we report their results on the synthetic dataset in Appendix F.

In this dataset, GCGM excels over other SSL methods in both \textit{Intersection} and \textit{Unfiltered} settings. It is worth noting that the performances of SCGM and GANN-GM significantly trail behind GCGM, widening the gap observed on the visual datasets. The reason is that the synthetic dataset does not require visual backbone processing or visual input, making SCGM and GANN-GM unable to exploit the visual aspects in their original design. Hence, they tend to perform poorly in general graph matching outside the visual domain. In contrast, our GCGM is graph-centric and does not rely on visual information, lending to its effectiveness and robustness in general graph matching.

\subsection{Ablation and Model Analyses}
\paragraph{Effect of Graph Augmentations}
\begin{table}[tbp]
    \centering
    \addtolength{\tabcolsep}{-1.5mm}
    \resizebox{\linewidth}{!}{
    \begin{tabular}{@{}c|cc|c|cc|cc@{}}
    \toprule
    \multirow{2}*{\shortstack[c]{Augmentation\\Set}} &\multicolumn{2}{c|}{Pascal VOC} & Willow & \multicolumn{2}{c|}{SPair-71k} & \multicolumn{2}{c}{Synthetic}\\
     & Intsec & Unfilt & Intsec & Intsec & Unfilt & Intsec & Unfilt\\
     \midrule\midrule
     $\mathcal{T}\setminus$NI & 56.9 & 36.6 & 94.8 & 61.8 & 34.9 & 57.9 & \textbf{40.5}\\
     $\mathcal{T}\setminus$NR &56.5 & 36.5 & \textbf{95.1} & 61.8 & 34.4 & 57.8 & 40.0\\
     $\mathcal{T}\setminus$ER & 57.3 & 37.2 & 95.0 & 59.8 & 32.5 & 57.9 & 40.0\\
     $\mathcal{T}\setminus$FS & \textbf{57.5} & 37.2 & 95.0 & 62.1 & 35.1 & 57.8 & 40.3\\\midrule
     $\mathcal{T}$ & 57.3 & \textbf{37.4}& 95.0 & \textbf{62.6} & \textbf{35.4} & \textbf{58.1} & 39.9\\
    \bottomrule
    \end{tabular}}
    \caption{Ablation study on graph augmentations.}
    \label{table:augmentaions}
\end{table}

We explored the contribution of different types of graph augmentation, by excluding each type from the augmentation set $\mathcal{T}$. As observed in Tab.~\ref{table:augmentaions}, Node Insertion (NI) and Node Replacement (NR) generally have a significant impact on the performance, particularly on Pascal VOC and SPair-71k. For the synthetic dataset, the two augmentation types appear not useful. A potential reason is that the node features are generated randomly, which means the feature differences between outliers and inliers are less significant. Thus, NI and NR can help less on the synthetic dataset. On the other hand, excluding Edge Removal (ER) and Feature Scaling (FS) generally results in a slight performance dip. In particular, ER appears to be the most useful to SPair-71k, due to the significant variability in viewpoint and scale in this dataset, as compared to other datasets \cite{min2019spair}. This variability leads to diverse geometric configurations of keypoints across images, which impacts and alters the graph structures formed. By randomly removing edges, ER encourages the model to focus on invariant features over variable graph structures, which is important on SPair-71k.

\paragraph{Effect of BiAS and Augmentation Pool} Tab.~\ref{table:fine-tune} compares the performance of GCGM under different strategies for selecting augmentations.
In the first category including `Random', `Tuning' and `Tuning + BiAS', we only adopt one hyperparameter setting for each augmentation type. In `Random', the hyperparameters of an augmentation type are initialized randomly, whereas in `Tuning' or `Tuning + BiAS', they are selected via 100 trials of TPE based on validation performance. Furthermore, `Random' and `Tuning' applies a uniform sampler when selecting the augmentation pair in each mini-batch, whereas `Tuning + BiAS' applies our BiAS sampler. In the second category including `Uniform' and `BiAS', we do not rely on just one set of hyperparameters for each augmentation type. As discussed in Sect.~\ref{sec:comprehensive_graph_augmentations}, we instantiate a family of augmentations from each type with different hyperparameter settings. In each mini-batch, we select an augmentation pair from all pairs of instantiations using either the uniform or BiAS sampler. 

In terms of performance, we observe that `Tuning' generally outperforms `Random', implying that hyperparameter settings are important to each augmentation type. Moreover, `Tuning + BiAS' generally improves over `Tuning', and `BiAS' consistently improves over `Uniform', highlighting the effectiveness of the BiAS sampler. Furthermore, using a large pool of augmentation pairs can compensate for the lack of hyperparameter tuning, as evident from the improvements made by `Uniform' and `BiAS' w.r.t.~`Random'. 

In terms of time cost, `Tuning' and `Tuning + BiAS' fare poorly due to the need of hyperparameter tuning. Yet, their performance rarely surpasses BiAS. On the other hand, the time cost of BiAS is two orders of magnitude smaller than tuning-based methods, and simultaneously achieves the best or near-best performance. It is also worth-noting that the weight updating strategy in BiAS incurs a negligible overhead, taking similar training time as `Random' or `Uniform' while achieving better results. 

In summary, BiAS offers a robust solution for selecting optimal augmentations, yet without the significant time overhead required by the tuning-based methods.

\begin{table}[t]
    \centering
    \addtolength{\tabcolsep}{-1.5mm}
    \resizebox{\linewidth}{!}{
    \begin{tabular}{@{}l|c|ccc|cc|ccc@{}}
    \toprule
    \multirow{2}*{Settings} & \multirow{2}*{$\mathcal{P}$} & \multicolumn{3}{c|}{Pascal VOC} & \multicolumn{2}{c|}{Willow} & \multicolumn{3}{c}{SPair-71k} \\
     & & Intsec & Unfilt & Time/h & Intsec & Time/h & Intsec & Unfilt & Time/h\\
     \midrule\midrule
     Random & \xmark & 55.0 & 35.9 & 0.26 & 93.8 & 0.04 & 61.4 & 35.2 & 0.38\\
     Tuning & \xmark & 55.9 & 36.8 & 23.76 & 94.8 & 4.54 & 61.5 & 35.6 & 31.96 \\
     Tuning + BiAS & \xmark &55.8 & 37.0& 23.95 & \textbf{95.4} & 4.54 & 61.9 & \textbf{36.0} & 31.88 \\
     \midrule
     \method{Uniform} & \cmark & 56.9 & 36.7 & 0.32 & 94.7 & 0.05 & 62.0 & 34.8 & 0.35 \\
     \method{BiAS} & \cmark & \textbf{57.3} & \textbf{37.4} & 0.39 & 95.0 & 0.05 & \textbf{62.6} & 35.4 & 0.34\\
    \bottomrule
    \end{tabular}}
    \caption{Performance of different initialization of augmentations and the use of augmentation pool. Time (hour) represents the total wall clock time spent on tuning the augmentations (for `Tuning' methods) and training the model. The `$\mathcal{P}$' column indicates if a pool of augmentation pairs is used.}
    \label{table:fine-tune}
\end{table}

\paragraph{Additional Model Analyses} Additionally, we experimented our framework with varying sizes of the augmentation pool and different combinations of the hyperparameter $\lambda$ and $\alpha$. We also investigated the effect of the individual design elements in BiAS, namely, momentum update and performance criterion. Finally, we compared GCGM with SCGM when the level of input information varies, to show the robustness of our approach given minimum input. Due to space constraint, we omit these results here, and present
them in Appendix F.

\section{Conclusion}
We introduced the Graph-centric Contrastive framework for Graph Matching (GCGM), a novel framework toward general graph matching without the need of any side information. GCGM utilizes a comprehensive pool of graph augmentations for self-supervised contrastive learning, which enhances matching robustness in the absence of side information. It is further complemented with a Boosting-inspired Adaptive Augmentation Sampler (BiAS), which dynamically selects the challenging augmentations for optimal results without tuning the hyperparameters associated with the augmentations. Together, through our experiments GCGM with BiAS achieves superior performance in graph matching, surpassing state-of-the-art self-supervised methods across various domains. At the same time, it is significantly more efficient than conventional augmentations with hyperparameter tuning.

\clearpage

\section*{Acknowledgements}
This research project is supported by the Ministry of Education, Singapore, under its Academic Research Fund Tier 2 (Proposal ID: T2EP20122-0041). Any opinions, findings and conclusions or recommendations expressed in this material are those of the author(s) and do not reflect the views of the Ministry of Education, Singapore.

\bibliographystyle{named}
\bibliography{ijcai24}

\end{document}